\documentclass[conference]{IEEEtran}
\ifCLASSINFOpdf
  \usepackage[pdftex]{graphicx}
\else
\fi
\usepackage[cmex10]{amsmath}
\DeclareMathOperator*{\argmax}{arg\,max}

\usepackage[tight,footnotesize]{subfigure}
\begin{document}
\title{Sensory Anticipation of Optical Flow\\ in Mobile Robotics}

%\author{\IEEEauthorblockN{Arturo Ribes}
%\IEEEauthorblockA{IIIA-CSIC - Campus UAB,\\
%             08290 Bellaterra, Barcelona, SPAIN\\
%Email: aribes@iiia.csic.es}
%\and
%\IEEEauthorblockN{Jes\'{u}s Cerquides}
%\IEEEauthorblockA{IIIA-CSIC - Campus UAB,\\
%             08290 Bellaterra, Barcelona, SPAIN\\
%Email: cerquide@iiia.csic.es}
%\and
%\IEEEauthorblockN{Ram\'{o}n L\'{o}pez de M\'{a}ntaras}
%\IEEEauthorblockA{IIIA-CSIC - Campus UAB,\\
%             08290 Bellaterra, Barcelona, SPAIN\\
%Email: mantaras@iiia.csic.es}}

\author{\IEEEauthorblockN{Arturo Ribes\IEEEauthorrefmark{1}\IEEEauthorrefmark{2},
Jes\'{u}s Cerquides\IEEEauthorrefmark{1},
Yiannis Demiris\IEEEauthorrefmark{2} and
Ram\'{o}n L\'{o}pez de M\'{a}ntaras\IEEEauthorrefmark{1}}
\IEEEauthorblockA{\IEEEauthorrefmark{1}IIIA-CSIC - Campus UAB, 
             08290 Bellaterra, Barcelona, SPAIN\\
Email: \{aribes, cerquide, mantaras\}@iiia.csic.es}
\IEEEauthorblockA{\IEEEauthorrefmark{2}Department of Electrical and Electronic Engineering, 
Imperial College London, SW7 2BT, UK\\
Email: y.demiris@imperial.ac.uk}}


%Telephone: (800) 555--1212, Fax: (888) 555--1212}


\maketitle

\begin{abstract}

%In the field of robotics, optical flow is a powerful cue that expresses how the 
%scene is moving w.r.t. the robot, encoding its geometry. As it is invariant to 
%scene appearance, it can be used in navigation tasks.
In order to anticipate dangerous events, like a collision, an agent needs to make long-term predictions. However, those are challenging due to uncertainties in internal and external variables and environment dynamics.
%In this work, we analyse how optical flow evolves over time, showing the importance of taking advantage of action awareness. 
A sensorimotor model is acquired online by the mobile robot using a state-of-the-art method that learns the optical flow distribution in images, both in space and time.
The learnt model is used to anticipate the optical flow up 
to a given time horizon and to predict an imminent collision by using 
reinforcement learning.
We demonstrate that multi-modal predictions reduce to simpler distributions once actions are taken into account. 

\end{abstract}



% \fi
\IEEEpeerreviewmaketitle

%%%%%%%%%%%%%%%%%%%%%%%%%%%%%%%%%%%%%%%%%%%%%%%%%%%%%%%%%%%%%%%%%%%%%%%%%%%%%%%%
\section{INTRODUCTION}

One of the objectives of developmental robotics is to autonomously learn the consequences 
of actions by interacting with the environment \cite{lungarella2003developmental}\cite{dearden2005learning}. By consequences, we denote the perceived effects in the agent's sensors. Acquired knowledge is
dependent on the sensorimotor capabilities of the agent and its own experience.

%Instead of equipping the robot with lasers or sonars, which provide valuable information for
%navigation tasks, 
%We advocate for the use of vision to extract information relative to navigation tasks.
%We made this decision because most of the image processing can be reused
%in other tasks that cannot be accurately solved with other sensors, like object recognition.

%Visual proprioception is the awareness of one's own movement and resulting postural adjustments produced by patterns of optical flow \cite{jjg1979affordance}.
Optical flow is very important for locomotion, providing information to the agent about how the scene is moving \cite{jjg1979affordance}\cite{warren1998visually}. The movement may be due to its own body motion or other objects moving around. 
It thus encodes the geometry and dynamics of the scene, and is invariant to appearance information. 

We can benefit from the fact that an agent is aware of the actions it performs, so it may learn
a forward model of how optical flow changes when it performs an action and use it 
%as a prior to help in the estimation or 
to capture task-relevant information like an imminent collision.

From a developmental perspective, as the early development of navigation is more related
to the dorsal pathway in primate vision, also referred as vision-for-action \cite{milner2008two}, 
that mainly deals with geometric and motion cues.
Doing so, we mitigate the effects of the high variability of scene or object appearance.

%However, optical flow estimation is a hard task and many methods have been developed
%for more than twenty years \cite{fleet2005optical}. Those methods can be divided in two %families. Local methods, such as 
%Lucas-Kanade algorithm \cite{lucas1981iterative}, are very fast but lack global consistency and can only measure the component of
%optical flow in the direction of image intensity gradient, 
%so it is difficult to get accurate estimates for the whole image.\\
%On the other side, global methods like \cite{horn1981determining}, minimize an energy functional that gives accurate dense
%fields of optical flow, but those are very slow to run in a mobile robot.

Although newborns can discriminate changes in heading with optical flow alone \cite{gilmore2004stability}, those are very primitive and need locomotor experience to further develop \cite{uchiyama2008locomotor}. There is also evidence of those visuo-motor couplings in 3-day old babies, which have positive feedback structures that modulate stepping behaviour \cite{barbu2009neonatal}. 

In this paper we study the mechanisms that enable an active agent to make long-term predictions of optical flow with a model that is learned dynamically. We analyse the optical flow distribution in terms of space and time, 
that is, what are the experienced optical flow values and how do they change in time. We show how complex the 
posterior distributions become when long-term predictions are needed, which breaks time-consistency assumption. The choice of 
one predictor or another should be made in terms of how the data is distributed.
Moreover, we use a generic state-of-the-art incremental online
learning algorithm \cite{heinen2010incremental} for the task of building a model to predict the optical flow perceived by a mobile robot.
Finally, as an application, the model is also used to learn a simple predictor for anticipating an imminent collision.

%The remainder of this paper is structured as follows. In Section 3, we explain our 
%proposed method for learning a predictive model of optical flow and how to apply it
%for increasing the effectiveness of the local estimation method. In Section 4, we
%give the details of our experimental setup, as well as the results obtained.
%Finally, in Section 5, we state our conclusions and future work in this research line.

%- Why is important to use optical flow in robotics?\\
%- Problems of optical flow estimation: the necessity of extra knowledge.\\
%- Related work applying OF in robotics and real-time approaches. \\
%- Overview of our approach.\\
%- Structure of the paper.\\

%%%%%%%%%%%%%%%%%%%%%%%%%%%%%%%%%%%%%%%%%%%%%%%%%%%%%%%%%%%%%%%%%%%%%%%%%%%%%%%%
\section{RELATED WORK}

Research in forward model learning and sensorimotor anticipation revolves around two main axis: length of predictions and direct applications of forward models. 

In our work we are very interested in providing long-term predictions. One option is to learn a model based on a differential equation of how sensor values change \cite{fujarewicz2007predictive}. Then we can anticipate sensory states at arbitrary times by simulating such a system, although accuracy decreases quickly depending on model complexity. Unfortunately, this cannot be reused directly to predict collisions and cannot handle multi-modality unless using a mixture. 
%The model presented in this work handles this naturally. 

In order to provide the agent with longer-term predictions, some authors proposed chaining forward models, where each one provides one-step predictions \cite{tani1999learning}\cite{gross1999generative}\cite{ziemke2005internal}\cite{hoffmann2007perception}. Their results showed that agents that anticipate sensory consequences of their actions behave more effectively than reactive agents. 
However, due to the intrinsic complexity in sensor data, some authors used a Mixture of Experts, where each expert was a Recurrent Neural Network (RNN) \cite{tani1999learning}. Experiments were conducted in simulated environments with low-dimensional sensor data, where it is not clear how well it could scale in more realistic environments.
Furthermore, this chaining process leads to accumulation of prediction errors, so authors proposed filtering schema based in PCA \cite{hoffmann2007perception} or using RNNs that also take as input the hidden state of the network from last step \cite{ziemke2005internal}. 

From the application point of view, many works use forward models to solve certain navigation related tasks. Forward models have been applied to generate expectations of sensory values, which have been used to correct noisy optical flow fields \cite{stephan2001neural} or to detect useful landmarks for navigation \cite{fleischer2003sensory}. If the forward model was acquired in an obstacle free environment, comparing expectations to novel sensory data also has been applied to detect obstacles \cite{nakamura1995motion}.
All those expectation-driven mechanisms could benefit from an incremental model as the one presented in this work to generate such expectations.

\section{METHODOLOGY}

When an agent is situated in an unknown environment, one of the first capabilities
that it needs to acquire is that of navigation, a task which purely relies in the geometric distribution
of objects in the agent's surroundings.

Among the many methods to extract the environment structure, we have selected optical
flow because it aggregates both spatial and dynamic information, which can be used to 
infer both the geometry and how things are moving, enabling the robot
to predict where are the obstacles located and time to collision. We use a GPU implementation of phase-based optical flow \cite{pauwels2011comparison}, which provides a dense flow field in real time.

%Local methods are suitable for real-time optical flow estimation. 
%In an early phase of our experiments, we
%used a fast implementation of the Lucas-Kanade method, available in the OpenCV library.
%However, this method provides noisy estimates and in large homogeneous areas the flow cannot be reliably estimated.
%After extensive experimentation, we decided to use the phase-based optical flow method from \cite{pauwels2011comparison}.
%This technique is more robust than Lucas-Kanade and provides dense flow fields, but CPU implementations are 
%very slow to run in real-time. Taking advantage of GPUs processing power, the authors in \cite{pauwels2011comparison} manage to 
%get enough frame rate to perform real-time processing.

The sensorimotor capabilities of our robot are defined as follow. The optical flow is computed at 
locations distributed on a uniform grid of $N$ by $M$. As it is a field of 2-D vectors, its 
dimensionality is $2NM$. We denote the optical flow at time $t$ using the random variable $OF_t$.
The robot also has access to proprioceptive data, in our case encoded as the linear and angular
velocities. The perceived velocity at time $t$ is extracted using the wheel encoders and denoted by the random variable $V_t$.
The action performed at time $t$ is defined as the desired linear and angular velocity and 
is captured by the random variable $A_t$.

The goal of the system is to anticipate what will be the perceived optical flow at $T$ time steps in the future,
having observed the current perceptions and the action we are performing. 
%This implies that we need to 
%learn a mapping between an input domain $X$ and output domain $Y$.

\subsection{Analysis of optical flow distribution}

\begin{figure*}[t]
\begin{center}
  \subfigure[Regions selected from $P(OF_t)$.]{\label{fig:cd_regions}\includegraphics[width=0.29\textwidth]{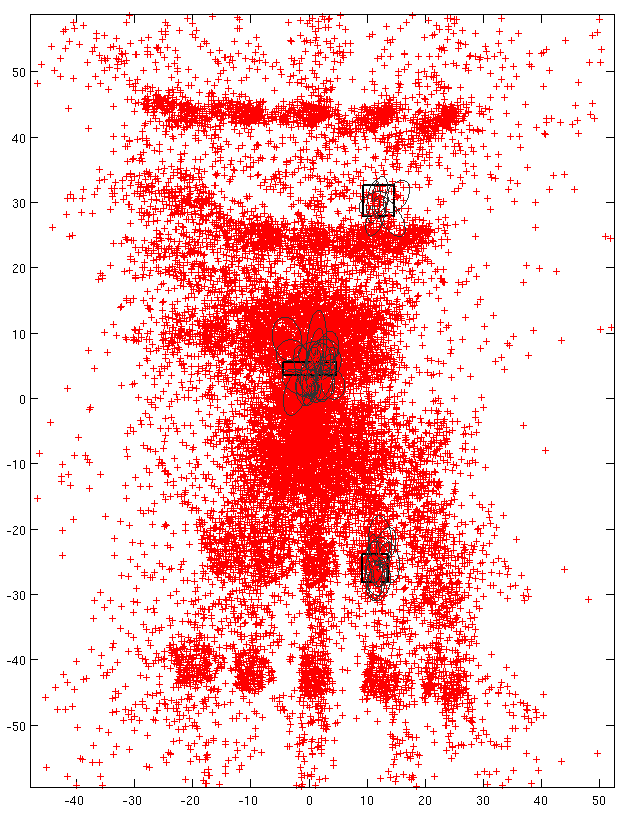}}        
  \subfigure[Conditional flow distributions for the selected regions $P(OF_{t+T} | OF_t)$.]{\label{fig:cd_timedelays}\includegraphics[width=0.7\textwidth]{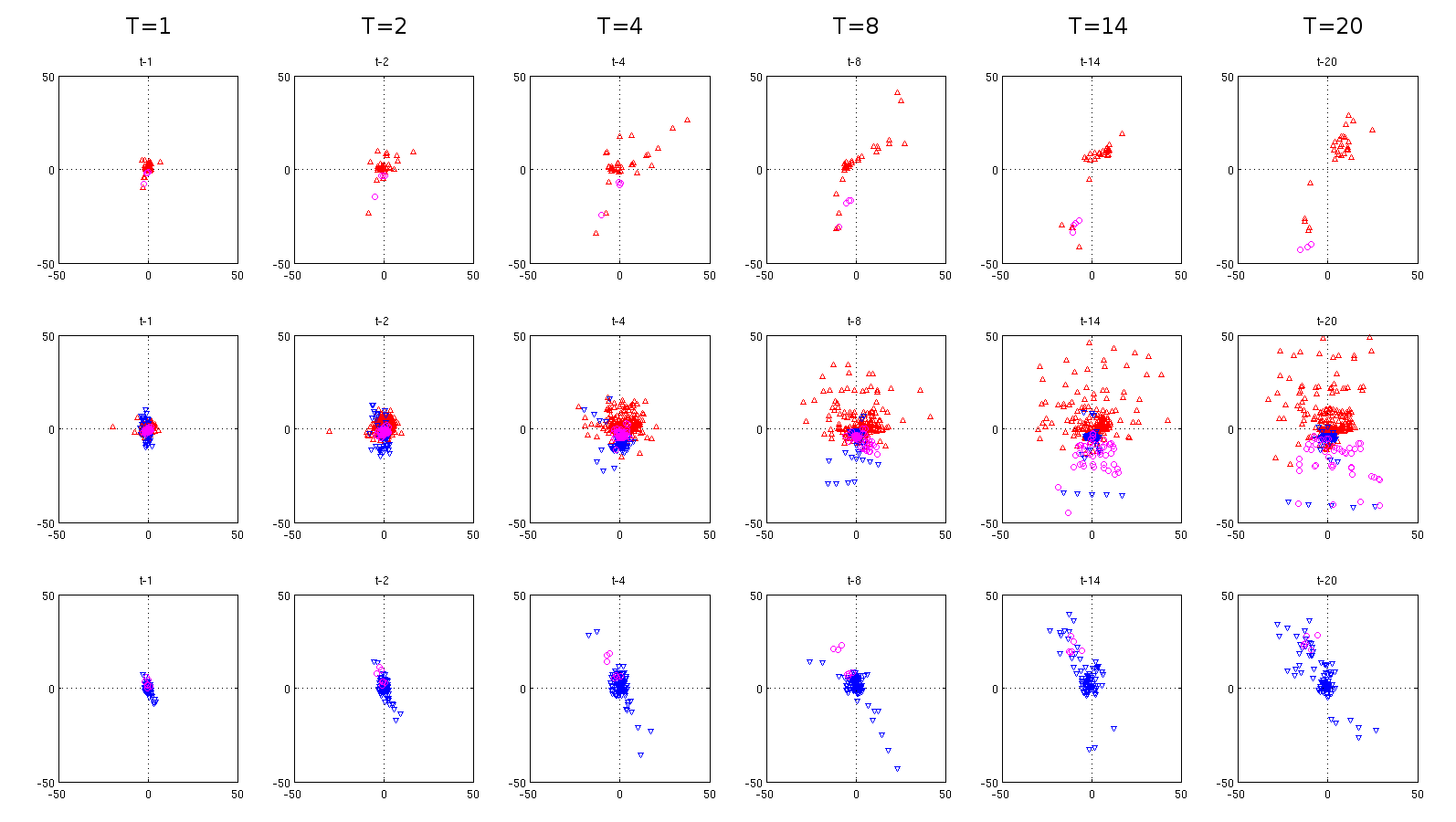}}
  \caption{Plot of the conditional distribution $P(OF_{t+T} | OF_t)$. In (a) a distribution of optical flow values $OF_t$ is depicted. Axes are flow in $X$ and $Y$ directions (pixels/sec). Each point represents an observed optical flow value. The big area in the middle shows that most of the time, small optical flows are observed, while the clusters in top and bottom of the image represent the optical flows when the robot moves forward/backward, present mainly in the bottom of the image, which moves faster. Small clusters can be identified due to the low spatial resolution used, as we sampled the optical flow in a grid of $5\times4$. In (b) the conditional distributions $P(OF_{t+T} | OF_t)$ are plotted, one row for each one of the selected regions, marked in (a) as black rectangles, and one column for different prediction horizons $T$. Action (forward/backward/stop) is encoded in different colour and shape. Axes represent the change in optical flow in $X$ and $Y$ directions, $\Delta OF_t = (OF_t - OF_{t-T})$.}
  \label{fig:conditional_distr}
\end{center}
\end{figure*}

Our initial hypothesis was that for a very small prediction horizon $T$, the change in optical flow is rather small,
so a na\"{i}ve predictor that assumes flow constancy in time would be enough for the task. 
We decided to analyse the data distribution to see which kind of predictors could be used for this task. Actually,
we were interested in the distribution $P(OF_t)$, looking for possible clusters or modalities, and how compact and
sparse they were. Figure~\ref{fig:cd_regions} shows the data distribution $P(OF_t)$ obtained by moving the robot forward and backward in our lab. \\
After identifying some modalities in the data, we were also interested in the distribution we need to use
to make predictions, $P(OF_t|OF_{t-T})$. Specifically, we looked for distributions that presented some multi-modality, which could indicate that changes in optical flow are due to an external factor, which we hypothesized as being the action $A_t$. Figure~\ref{fig:conditional_distr} shows the distribution $P(OF_t|OF_{t-T})$ for some regions in $OF_{t-T}$. 
%At this point, we need to clarify that we experimented with two ways of predicting optical flow. 
%One is predicting the actual optical flow that will be observed $OF_t$, and the other is to predict 
%the change in the flow vectors $\Delta OF_t = (OF_t - OF_{t-T})$. 
%As both approaches have its advantages, we discuss them in the results section.

\begin{figure}
\begin{center}
\includegraphics[width=0.40\textwidth]{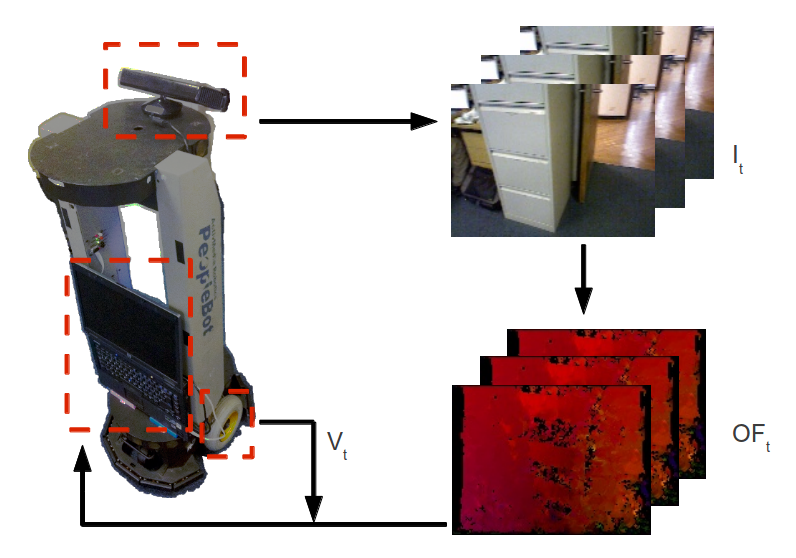} 
\caption{Pioneer PeopleBot with a mounted Kinect providing images $I_t$, which are processed to obtain optical flow $OF_t$, our visual input. Proprioception sensors provide wheel velocities $V_t$ and everything is processed in the laptop.}
\label{fig:sensor_flow}
\end{center}
\vspace{-2.0em}
\end{figure}

The analysis showed that we needed a method that provides a model which is learnt quickly and is useful after a short period of time, i.e. an incremental and on-line method. We propose to learn the joint distribution of current optical flow ($OF_{t}$) and the previous 
action ($A_{t-T}$), proprioception ($V_{t-T}$), and optical flow ($OF_{t-T}$) and use it as a forward model in prediction. Figure \ref{fig:sensor_flow} shows the robot used in our experiments and how sensor information flows through the system. An example image and resulting optical flow shows the kind of untextured structured environment where the robot navigates.\\

\subsection{Definition of our model}

The main problem with learning a distribution like the one described above is its dimensionality and the need for marginalizing over some variables to turn the joint distribution into a conditional one for making predictions. We decided to make some assumptions to lower the complexity of the resulting approach, as we need the whole system to run in real time.

The first assumption made is a Markovian one, stating that $OF_t$ is conditionally independent, given $OF_{t-T}, A_{t-T}, V_{t-T}$, of $OF_{t-i}, A_{t-i}, V_{t-i}$ s.t. $i \in [1, \infty) \cap \lbrace T \rbrace$. That assumption, although fairly strong, greatly reduces the model complexity while providing a model which still has some short-term memory.

In order to ease the notation, we define $X$ as the set of 
input variables, $X = \lbrace OF_{t-T}, A_{t-T}, V_{t-T} \rbrace$ and $Y$ is the set of output variables, $Y = \lbrace OF_{t} \rbrace$.

The second assumption is that the distribution can be approximated using a Gaussian Mixture Model $M$. 
The method chosen to learn it is an incremental version of multivariate GMM \cite{heinen2010incremental}. By feeding the algorithm with the data samples as they arrive from the sensors, this method 
learns while the robot is moving, and as it is incremental, after a few seconds gives good predictions 
for common situations, e.g. wandering around with no obstacles.\\
This method also allocates new clusters to the mixture when there is a low likelihood that
the current model explains the new sample. The only parameters to choose are the threshold on
the mixture component likelihood and the initial covariance matrix for initializing new components.

With the aim of easing the prediction of optical flow, we made another conditional independence assumption,
treating $Y$ as conditionally independent of $X$, given the mixture $M$. 
This assumption implies that each multivariate Gaussian component $m_j$ has two separate mean vectors and covariance matrices for each set of independent variables, that is $\mu^X_j$, $\Sigma^X_j$, $\mu^Y_j$ and $\Sigma^Y_j$.

\subsection{Alignment of sensory streams}

\begin{figure}[t]
\begin{center}
\includegraphics[width=0.48\textwidth]{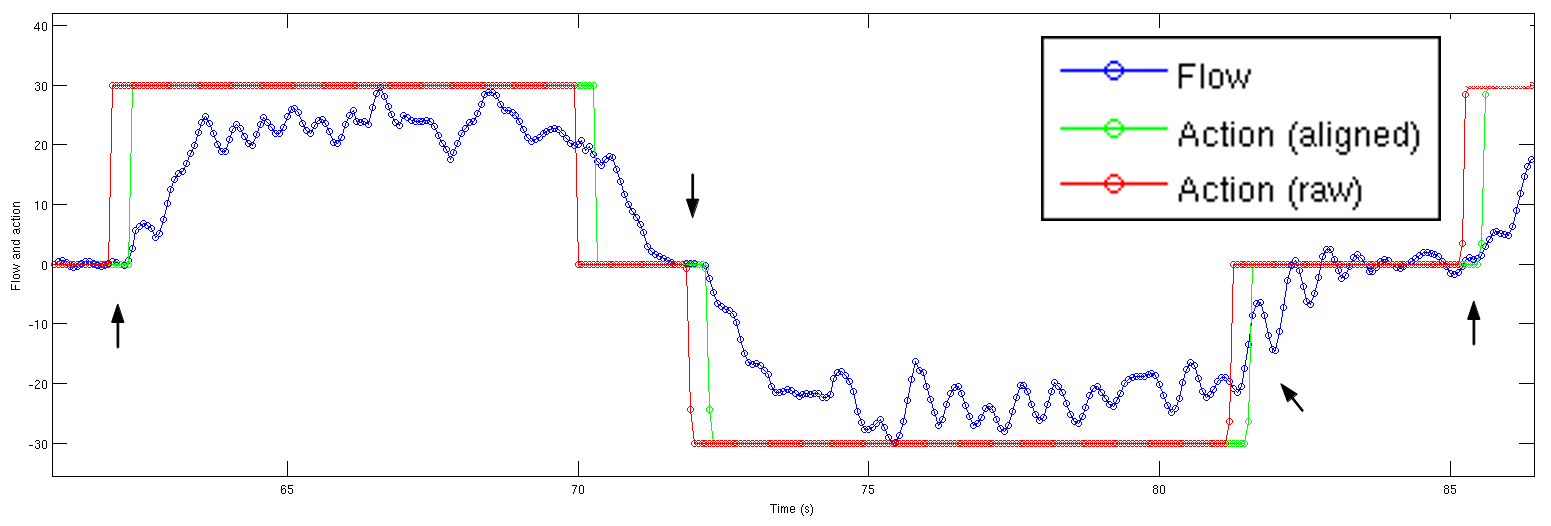} 
\caption{Alignment of the optical flow stream to the action stream. Horizontal and vertical axes are time and vertical optical flow, respectively. The step signals are the aligned and unaligned action, scaled for visualisation purposes. It can be appreciated how changes in the aligned action, indicated by arrows, are more correlated with changes in optical flow.}
\label{fig:of_alignment}
\end{center}
\end{figure}

The use of time-series coming from different sensors has an associated issue that needs to be addressed first. 
As it happens with animals, signals from different senses arrive at slightly different timings, so the brain needs to align those signals to 
extract more information. In our system, we may observe this when we issue an action command $a_t$ and, due to the physical 
characteristics of the robot, we do not capture the effects in the visual sensors until some time later.\\
In order to model this time delay between signals from different modalities, we followed a methodology in the fashion of \cite{dearden2005learning}, taking as the optimal time-delay as the one that maximizes the log-likelihood of the data given the model parameter.
%which consists in training N separate models, each one introducing a different time-delay in the signal to be aligned, in our case the action $A_t$.
%Then we plot the log-likelihood of the data and take the model that maximizes it after the model seems to converge.\\
%As a validation test, we aligned by hand some sequences of the data to check if the estimated time-delay is correct. 
In Figure \ref{fig:of_alignment} we show the alignment of action signal using the time-delay estimated in our experiments, which is the same we obtained manually.

\subsection{Learning and prediction using the GMM}

%Having the signals aligned, now we focus on learning the optical flow distribution described above to use it as a predictor. 
Basically the GMM can be visualized as a kernel density estimator if we
set the number of components equal to the number of data samples. As we reduce the number of components, the GMM represents a compressed
dataset that approximates the underlying data distribution. It is desirable to have a trade-off between compression and representativeness,
as it affects both to prediction accuracy and real-time performance of the algorithm.

As described by \cite{heinen2010incremental}, both the learning algorithm and prediction algorithm compute the likelihoods of hundreds of multivariate normal distributions. 
%We noticed that in the case of prediction, not all the model's components need to be used. 
In our case, we set a threshold on the minimum mass that a component needs to incorporate in order to be used as predictor, so very young components or spurious ones are not used.
However, learning does compute likelihoods for every component, as it is necessary for computing posterior probabilities.
%, so we expect to introduce a pruning mechanism in future work to keep model complexity manageable.

\cite{heinen2010incremental} show the update equations for the mixture components, which basically add a term to the mean and covariances, weighted by the proportion in which the sample's mass contributes to the mixture component. If this proportion is below a certain threshold, which we set to $10^{-4}$ in our experiments, we do not update the component. \\
This modification alleviates the cost of updating the mixture, given that each time we update the covariance matrix, we need to recompute its inverse and determinant to be able to evaluate the density function.

After the model is learnt, we can feed the sensor readings at the previous time step and 
obtain an estimate of what will be the optical flow in the next frame.
The optimal optical flow prediction $y^*$ is defined probabilistically as:
\begin{equation}
\label{eqn:opt_flow}
y^*(x) = \argmax_y P(Y=y|X=x)
\end{equation}
After applying the first assumption, i.e. introducing the mixture model $M$, and applying Bayes rule we have:
\begin{equation}
P(Y|X) = \sum_{M}P(Y|M)\frac{P(X|M)P(M)}{P(X)}
\end{equation}
As we are interested only in the MAP, we can drop the constant term $P(X)$, so the resulting equation is:
\begin{equation}
y^*(x) = \argmax_y \sum_{M}P(Y=y|M)P(X=x|M)P(M)
\end{equation}

In our case, we do this inference in two steps. First, we compute the most probable mixture component $m_{j^*}$ such
that $j^*(x) = \argmax_j P(m_j|X=x)$. After having identified the component, the posterior for $Y$ is given by the MAP of
the corresponding multivariate Gaussian, which is $\mu_{j^*}^X$. This is an approximation, as instead of the summation for
all the components, we take the component with maximum activation. 

\begin{figure}[t]
\begin{center}
\includegraphics[width=0.35\textwidth]{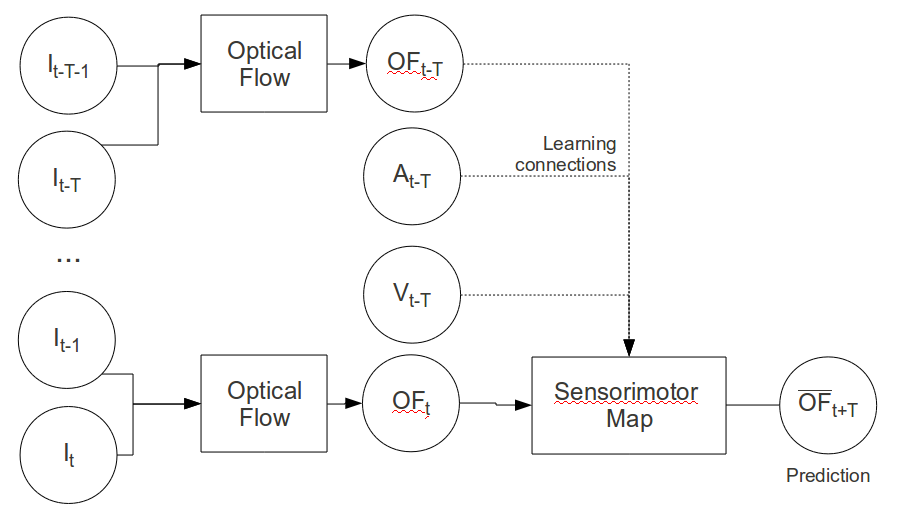} 
\caption{Diagram of the presented system. For learning, it takes samples from $(OF_{t-T}, A_{t-T}, V_{t-T}, OF_t)$.
For prediction, it uses $(OF_t, A_t, V_t)$ to predict $OF_{t+T}$.}
\label{fig:system_diagram}
\end{center}
\end{figure}

Figure \ref{fig:system_diagram} shows the proposed system. It depicts the connections between sensorimotor signals at time $t-T$
and time $t$ to learn the model, and the connections from $OF_t$ and $A_t$ and $V_t$, not shown in the image, to predict
optical flow at time $t+T$.

\subsection{Application: Anticipating a collision}

We designed an application to check if the mixture components capture enough information to be useful to anticipate the binary signal of the robot's bump sensors. 
That is, we check if it can predict an immediate collision. This application is very similar to that described by \cite{sutton2011horde}, where they use multiple predictors to anticipate sensor values of a robot.\\
Instead of introducing a new variable into the model, we treated the problem as temporal credit assignment. Each time the robot bumped into an object, we assigned credit for that bump to the components that were active in the last $N$ frames. We apply an exponential falloff depending on the time of activation and the discount factor, which is manually set. The value is added to an accumulator and used as the \textit{collision value} of the component, providing evidence for a collision in the near future.\\
Anticipation of a collision event is done as follows. First, the active mixture components are computed from the current optical flow values for each position in the sample grid. Then, the optical flow can be predicted and the collision value of the active components is averaged to output a collision signal. \\
The collision signal is highly correlated with a collision event likely to happen in the near future, which is around 2 seconds, depending on how big the obstacle is.

%%%%%%%%%%%%%%%%%%%%%%%%%%%%%%%%%%%%%%%%%%%%%%%%%%%%%%%%%%%%%%%%%%%%%%%%%%%%%%%%
\section{EXPERIMENTAL SETUP}

Our experiments are done using a Pioneer Peoplebot with a mounted Kinect camera. We have attached a laptop with a Core 2 Duo 1.8Ghz processor, 2GB of RAM and an NVIDIA Quadro 570M GPU where the optical flow is computed for 320x240 images.
No special arrangement of furniture or objects in the lab was done, with the aim of situating the robot in a realistic environment.
%Both robots have different physical and dynamical features that permit us to study our results 
%in terms of the scalability to different robotic platforms.
The robot is controlled using a joystick, so all the actions are performed by a human.
We decided not to use any action decision algorithm because we are concerned with the
learning capacity of our system, so we can drive it to challenging situations as
required in order to stress its acquired knowledge.

%We attached a computer to the robot, which sends commands to the robot's drivers,
%reads back proprioception data, computes optical flow from camera images and sends it
%all to the base station, where all the learning is done. We implemented the whole 
%system using ROS \cite{quigley2009ros}, which has lots of already implemented features and makes it 
%easy to setup a distributed system like ours.

%To test the parameters of the system, we recorded several sequences in both labs so we can see the effect of different parameters on the system.

The action space of the robot has been restricted to five actions: stop, forward, backward,
turn left and turn right, all at constant velocities fixed beforehand. In the experiments
reported here, we used $0.3 m/s$ for linear velocity and $0.6 rad/s$ for the angular velocity.

%As the model complexity is linear in the number of components of the model, we decided to 
%use no more than 50 flow samples per image, so we tested different resolutions, e.g. $4\times3, 6\times4, 8\times6$, to see which kind of events
%or objects the model learns to represent. For instance, with a very coarse resolution we could perceive the 
%approach of big obstacles, like a wall or a cabinet, but small objects like a box or
%a bottle in the floor, could not be detected.

In the case of prediction, we evaluated the mass distribution among components, and adjusted the mass threshold
to use at least 90\% of the model's mass. This usually corresponds to less than 10-15\% of the components, depending on how sparsely distributed the mixture components are.

%Learning was done by reproducing the sensorimotor stream of the sequences recorded with the mobile robot, and starting again depending on how much time we set for learning. Once the learning was done, we tested different mass thresholds to check the method accuracy depending on the model complexity.

The evaluation of the method was done by looking at two different measures. One is a common error measure in optical
flow estimation, the average end-point error (AEPE) between two flow fields. The other measure is a likelihood ratio, explained below. 
We also extracted the average angular 
error (AAE) but it is very unstable when flow magnitude is nearly zero, unless some parameter is introduced.

We do not have a ground truth for the sequences recorded, so, instead of analysing the AEPE in absolute terms, we normalize it by
the error that a na\"{i}ve predictor would do. This predictor is assumes a constant optical flow, i.e. $f(OF_t) = OF_{t-T}$, so basically we
should expect to do better in the discontinuities and with a high prediction horizon $T$.

We experimented with two ways of predicting optical flow. One is predicting the actual optical flow that will be observed $OF_t$, and the other is to predict 
the change in the flow vectors $\Delta OF_t = (OF_t - OF_{t-T})$. 
%Although the former representation provides some sort of discrete prediction of optical flow, which could be adequate in some setups, 
We chose the later because it gives better results and is more compatible for comparing with 
the na\"{i}ve predictor, which assumes that the time derivative of optical flow is zero.

Besides the approximation error, we were also interested in seeing how confident is the model in its predictions, 
as what we really anticipate is a distribution over possible flow values, and just take the MAP as the optimal predicted value.
However, the predicted distribution remains to be tested. It could happen that we get a high AEPE but that the likelihood of the 
predicted value was only a bit higher that the true value, so we should account for that in our results.
%This is why we also computed the log-likelihood that the observed optical flow fits the predicted flow distribution, which 
We show this as the log of the likelihood ratio between the na\"{i}ve predictor and the learnt model. 

We also decided to test separately if the introduction of the action $A_t$ in the model increases the quality of predictions or not.
Two different models were trained and compared, one that models $P(OF_t, OF_{t-T})$ and another that models $P(OF_t, OF_{t-T}, A_{t-T})$.
It should be noted that we did not include proprioception sensor information $V_t$ in this experiments, as we think that in the 
environments we test our robotic platform, the information provided will be highly redundant with that of the action.

%%%%%%%%%%%%%%%%%%%%%%%%%%%%%%%%%%%%%%%%%%%%%%%%%%%%%%%%%%%%%%%%%%%%%%%%%%%%%%%%
\section{RESULTS}

%\begin{figure}[t]
%\begin{center}
%\includegraphics[width=0.4\textwidth]{img/log_lhood_of.png} 
%\caption{Histogram of the optical flow distribution, in horizontal and vertical axis, extracted using a grid of $5\times4$. 
%Log-likelihood is encoded in colour. Note the clusters corresponding to different sensors.}
%\label{fig:of_hist}
%\end{center}
%\end{figure}

The optical flow distribution for all the sensors $P(OF_t)$, plotted in Figure \ref{fig:cd_regions}, with x and y axes being the horizontal and vertical flow values, respectively. 
%Colour encodes the log-likelihood of $OF_t$. 
The distribution presents clusters clearly defined for each row and column of sensors, with a big cluster in the center corresponding to the low flow values. \\
The conditional distribution $P(OF_t, A_{t-T} | OF_{t-T}=x)$ is shown in Figure \ref{fig:conditional_distr} for 3 different regions (black squares in Figure \ref{fig:cd_regions}) and for different time-delays $T$ (one row in \ref{fig:cd_timedelays} for each region and one column for each time-delay $T$).
Action is encoded in color and shape, corresponding to forward, backward and stop actions in the sequence depicted. 
From this plot we can see clearly why the constant predictor does better for small prediction horizons. That is, regardless of which region we condition on, 
we can see that for $T\leq2$, the conditional distribution is mostly uni-modal and centred in zero, so the constancy assumption of the na\"{i}ve predictor holds.
However, for predictions more than 10-15 time steps ahead, the distribution is more entropic, presents multiple modes that are not usually zero-centred and, most importantly, 
action information provides valuable information to segment the distribution into different modes.

The results of the alignment of the different sensorimotor streams are depicted in Figure \ref{fig:of_alignment}. As can be observed, 
the changes in the aligned action signal $A_{t-T}$ are more correlated with significant changes in the flow signal $OF_t$ than the unaligned action $A_t$. The best parameter was found to be $T=6$. 
%, obtained as the parameter that gives the maximum model likelihood.

%\begin{figure}[h]
%\begin{center}
%\includegraphics[width=0.4\textwidth]{img/alignment_likelihood.png} 
%\caption{Model likelihood for different values of parameter $T$ in the alignment of signal $A_{t-T}$ with the optical flow stream $OF_t$.}
%\label{fig:alignment_likelihood}
%\end{center}
%\end{figure}

\begin{figure}[h]
\begin{center}
\includegraphics[width=0.30\textwidth]{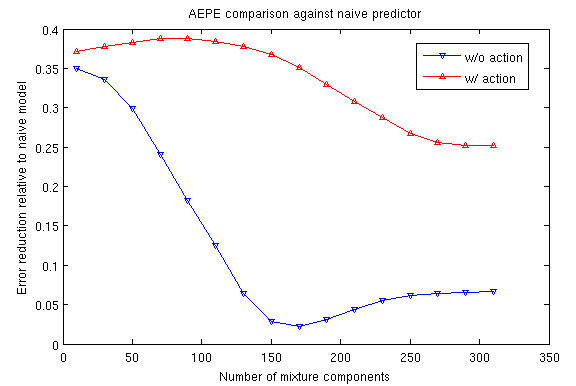} 
\caption{Relative AEPE error between na\"{i}ve predictor and GMM with and without action information. Taking into consideration action provides a model less sensitive to model complexity.}
\label{fig:aepe_numcomponents}
\end{center}
\end{figure}

Regarding the learning results, first we show the AEPE errors for different parameters of the system. Figure \ref{fig:aepe_numcomponents} shows 
the AEPE error as the percentage in error reduction relative to the na\"{i}ve predictor error, i.e. $e=1-\frac{err_{GMM}}{err_{naive}}$, plotted against the number of mixture components. We can see that predictions without 
using action information only reduce prediction error if we use compact models. However, after incorporating the action in our model, 
prediction error is robustly reduced by half, almost independently of the model density.

\vspace{-0.5em}
\begin{figure}[h]
\begin{center}
\includegraphics[width=0.30\textwidth]{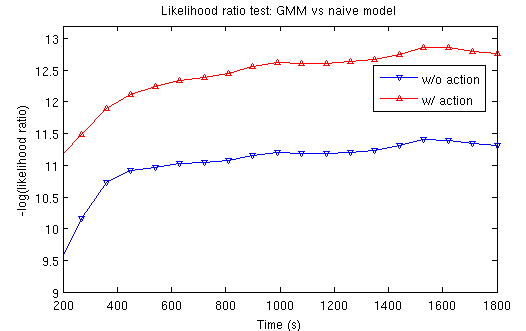} 
\caption{Likelihood ratio test between na\"{i}ve predictor and GMM with and without action information.}
\label{fig:lhood_ratio_test}
\end{center}
\end{figure}

\begin{figure*}[t]
\begin{center}
\includegraphics[width=1.0\textwidth]{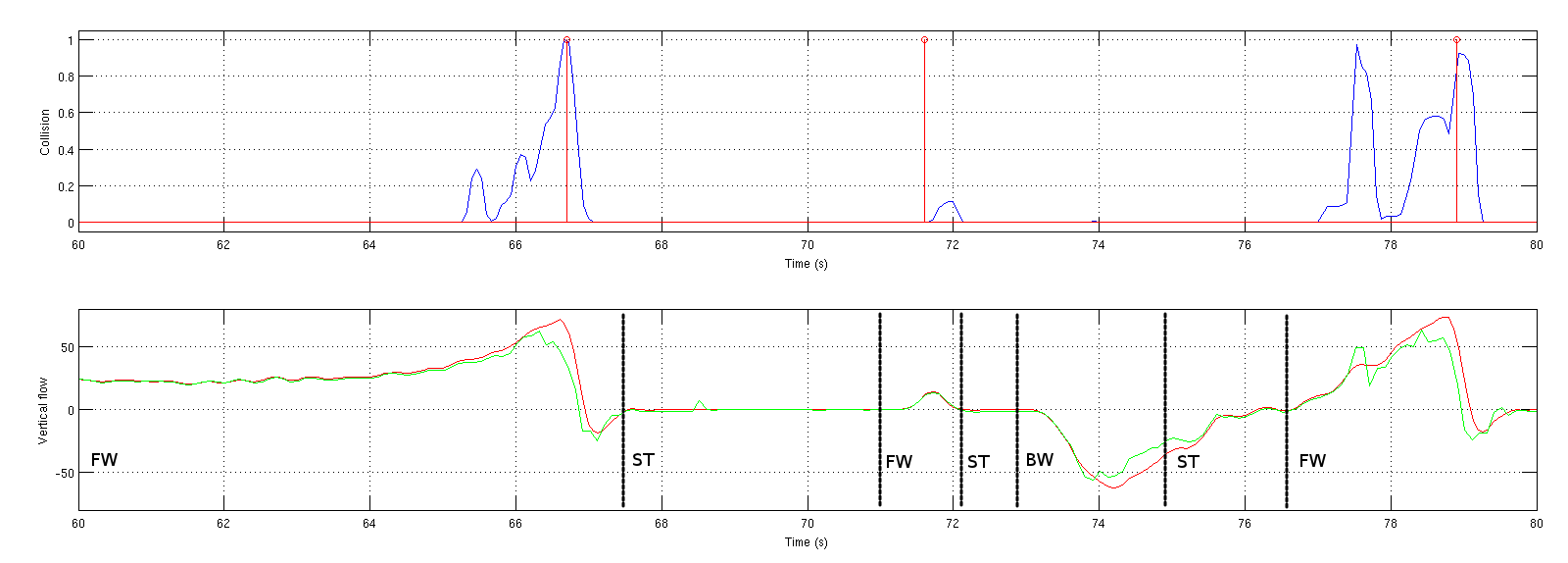} 
\caption{Plot of the sensorimotor signals in the collision anticipation experiment. On top we show collision signal, which is related to the value of the current state for predicting the event, learnt by reinforcement learning. On bottom we show the observed (red) and predicted (green) values of vertical flow. For visualisation purposes, the sequence is segmented using vertical bars when action changes. Actions are: forward (FW), stop (ST) and backward (BW).}
\label{fig:collisions}
\end{center}
\end{figure*}

We also computed the logarithm of the likelihood ratio between the na\"{i}ve predictor and the two versions of the GMM, with and without action
information. In Figure \ref{fig:lhood_ratio_test} we can see the results of this test, which indicate that our GMM model gives better predictions
than the na\"{i}ve model.

Next, we comment on the results of our model when applied to collision
anticipation.
After the model was bootstrapped by learning for some time, we
reproduced a sequence containing bumps into
an obstacle and the model quickly learned to anticipate
the collision up to 2 seconds before it happened, which
is a bit later than the time when the object fills a significant part of the field of view.

Results are depicted in Figure \ref{fig:collisions}. Both the collision prediction signal and the collision
events are plotted in the upper graph. It can be appreciated how the collision can be anticipated with a horizon above 1 second. The only collision which is not detected happens when the robot is touching the 
obstacle, so a forward action triggers the binary bumpers, but optical flow does not change significantly.
The middle graph shows the observed and predicted optical flows $OF_t,\widehat{OF_t}$, and action $A_t$ is plotted 
in the bottom graph.

%%%%%%%%%%%%%%%%%%%%%%%%%%%%%%%%%%%%%%%%%%%%%%%%%%%%%%%%%%%%%%%%%%%%%%%%%%%%%%%%
\section{CONCLUSIONS AND FUTURE WORK}

In this paper, we have presented a method to learn optical flow distribution when action and proprioception are observed, as is the case in the mobile robotics field.
%This differs from standard computer vision, where images are passively acquired and we have
%no information on what caused the perceived image. 
We show that taking advantage of action improves the results making predictions more robust.

When the task at hand is anticipating sensor values at a significantly high prediction 
horizon, our analysis of the optical flow dynamics provided evidence for rejecting
the flow time-constancy assumption. This called for the application of 
machine learning techniques to extract a representative model. 

We used the learnt model to accurately predict optical flow in advance, with a 
computation that can be done in real-time.

As an application of the model, we presented a collision anticipation mechanism that
builds on top of a learnt model and anticipates a collision when an object 
is approaching the robot.

We plan to apply this model to build an attention model. That will allow the prediction and estimation 
of optical flow to be interleaved in time. Also, we can use this
model as a joint observation and dynamics model in techniques like HMM or particle filtering.
We are also working in a principled extension to automatically delete spurious components and to refine the underlying structure.

%\section*{Acknowledgement}




%\bibliographystyle{IEEEtran}

\bibliographystyle{./IEEEtran}
\bibliography{ribes12}

\end{document}